\documentclass[runningheads]{llncs}
\usepackage{graphicx}
\usepackage{amsmath,amssymb} 
\usepackage{color}
\usepackage[width=122mm,left=12mm,paperwidth=146mm,height=193mm,top=12mm,paperheight=217mm]{geometry}

\usepackage{algorithm2e}
\usepackage{amsfonts}
\usepackage{algorithmicx}
\usepackage{algpseudocode}

\usepackage{times}
\usepackage{epsfig}
\usepackage{graphicx}
\usepackage{amsmath}
\usepackage{amssymb}

\usepackage{indentfirst}
\usepackage{algorithm2e}
\usepackage{amsfonts}
\usepackage{algorithmicx}
\usepackage{url}
\usepackage{algpseudocode}
\DeclareMathOperator*{\argmin}{arg\,min}
\def\1{\mbox{\bf 1}}
\def\A{{\bf A}}

\def\bmu{{\boldsymbol{\mu}}}

\def\B{{\bf B}}

\def\c{{\bf c}}
\def\D{{\bf D}}

\def\I{{\bf I}}
\def\J{{\bf J}}

\def\p{{\bf p}}

\def\Q{{\bf Q}}
\def\q{{\bf q}}
\def\R{{\bf R}}
\def\Real{\mathbb{R}}

\def\S{{\bf S}}
\def\s{{\bf s}}
\def\Sig{\boldsymbol{\Sigma}}
\def\U{{\bf U}}

\def\V{{\bf V}}
\def\x{{\bf x}}
\def\X{{\bf X}}
\def\y{{\bf y}}

\def\Y{{\bf Y}}

\def\T{\mathcal{T}}
\def\oO{\mathcal{O}}
\begin{document}
\pagestyle{headings}
\mainmatter

\def\ECCV16SubNumber{1763}  

\title{Cascaded Continuous Regression for Real-time Incremental Face Tracking} 

\titlerunning{Cascaded Continuous Regression for Real-time Incremental Face Tracking}

\authorrunning{E. S\'anchez-Lozano, B. Martinez, G. Tzimiropoulos and M. Valstar}

\author{Enrique S\'anchez-Lozano, Brais Martinez, \\ Georgios Tzimiropoulos, and Michel Valstar}
\institute{Computer Vision Laboratory. University of Nottingham \\ \email{\{psxes1,yorgos.tzimiropoulos,michel.valstar\}@nottingham.ac.uk}}

\maketitle

\begin{abstract}
\noindent This paper introduces a novel real-time algorithm for facial landmark tracking. Compared to detection, tracking has both additional challenges and opportunities. Arguably the most important aspect in this domain is updating a tracker's models as tracking progresses, also known as incremental (face) tracking. While this should result in more accurate localisation, how to do this online and in real time without causing a tracker to drift is still an important open research question. We address this question in the cascaded regression framework, the state-of-the-art approach for facial landmark localisation. Because incremental learning for cascaded regression is costly, we propose a much more efficient yet equally accurate alternative using continuous regression. More specifically, we first propose cascaded continuous regression (CCR) and show its accuracy is equivalent to the Supervised Descent Method. We then derive the incremental learning updates for CCR (iCCR) and show that it is an order of magnitude faster than standard incremental learning for cascaded regression, bringing the time required for the update from seconds down to a fraction of a second, thus enabling real-time tracking. Finally, we evaluate iCCR and show the importance of incremental learning in achieving state-of-the-art performance. Code for our iCCR is available from \url{http://www.cs.nott.ac.uk/~psxes1}
\end{abstract}

\section{Introduction}
\label{sec:introduction}

\noindent The detection of a sparse set of facial landmarks in still images has been a widely-studied problem within the computer vision community. Interestingly, many face analysis methods either systematically rely on video sequences (e.g., facial expression recognition \cite{dhall14}) or can benefit from them (e.g., face recognition \cite{zhou03}). It is thus surprising that facial landmark tracking has received much less attention in comparison. Our focus in this paper is on one of the most important problems in model-specific tracking, namely that of updating the tracker using previously tracked frames, also known as incremental (face) tracking.

The standard approach to face tracking is to use a facial landmark detection algorithm initialised on the landmarks detected at the previous frame. This exploits the fact that the face shape varies smoothly in videos of sufficiently high framerates: If the previous landmarks were detected with acceptable accuracy, then the initial shape will be close enough for the algorithm to converge to a ``good'' local optimum for the current frame too. Hence, tracking algorithms are more likely to produce highly accurate fitting results than detection algorithms that are initialised by the face detector bounding box. 

However, in this setting the tracker still employs a generic deformable model of the face built offline using a generic set of annotated facial images, which does not include the subject being tracked. It is well known that person-specific models are far more constrained and easier to fit than generic ones \cite{gross05}. Hence one important problem in tracking is how to improve the generic model used to track the first few frames into an increasingly person-specific one as more frames are tracked.

\begin{figure}[t!]
\begin{center}
\includegraphics[width=0.93\columnwidth]{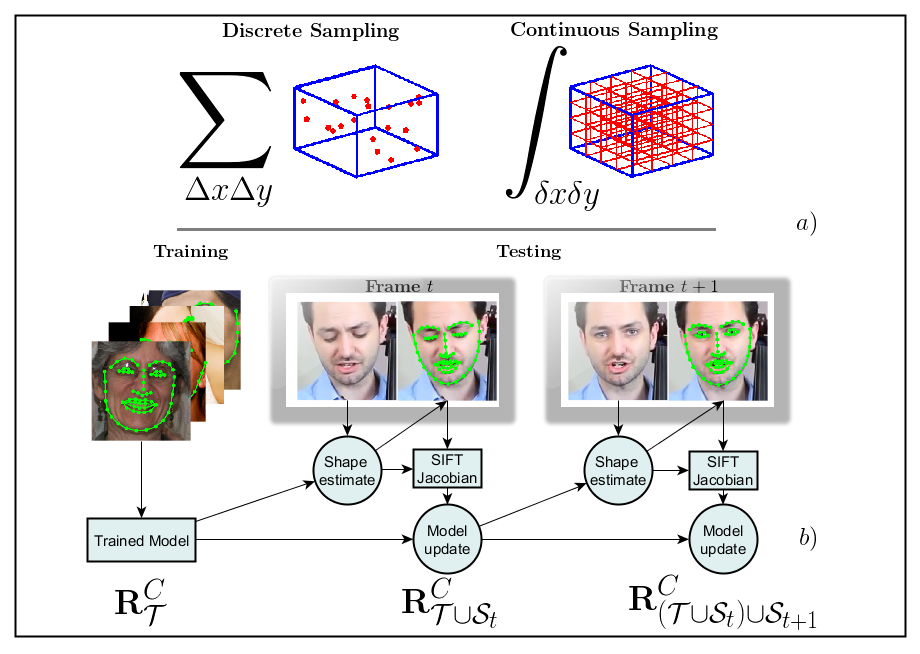}
\caption{Overview of our incremental cascaded continuous regression algorithm (iCCR). a) shows how continuous regression uses all data in a point's neighbourhood, whereas sampled regression uses a finite subset. b) shows how the originally model $\R_{\mathcal{T}}$ learned offline is updated with each new frame.}
\label{f:abstract}
\end{center}
\end{figure}

This problem can be addressed with incremental learning, which allows for the smart adaptation of pre-trained generic appearance models. Incremental learning is a common resource for generic tracking, being used in some of the state-of-the-art trackers \cite{hare16,wang15}, and incremental learning for face tracking is by no means a new concept, please see Ross et al. \cite{ross08} for early work on the topic. More recently, incremental learning within cascaded regression, the state-of-the-art approach for facial landmark localisation, was proposed by Xiong \& De la Torre \cite{xiong14} and independently by Asthana et al. \cite{asthana14}. However, in both \cite{xiong14} and \cite{asthana14} the model update is far from being sufficiently efficient to allow real-time tracking, with \cite{asthana14} mentioning that the model update requires 4.7 seconds per frame. Note that the actual tracking procedure (without the incremental update) is faster than 25 frames per second, clearly illustrating that the incremental update is the bottleneck impeding real-time tracking. 

If the model update cannot be carried out in real time, then incremental learning might not be the best option for face tracking - once the real-time constraint is broken in practice one would be better off creating person-specific models in a post-processing step \cite{sagonas14} (e.g., re-train the models once the whole video is tracked and then track again). That is to say, without the need and capacity for real-time processing, incremental learning is sub-optimal and of little use.

Our main contribution in this paper is to propose the first incremental learning framework for cascaded regression which allows real-time updating of the tracking model. To do this, we build upon the concept of continuous regression \cite{sanchez12} as opposed to standard sampling-based regression used in almost all prior work, including \cite{xiong14} and \cite{asthana14}. We note that while we tackle the facial landmark tracking problem, cascaded regression has also been applied to a wider range of problems such as pose estimation \cite{dollar10}, model-free tracking \cite{wang15} or object localisation \cite{yan14}, thus making our methodology of wider interest. We will release code for training and testing our algorithm for research purposes.

\subsection{Contributions}

\noindent Our main contributions are as follows:

\begin{itemize}
\item We propose a complete \textbf{new formulation for Continuous Regression}, of which the original continuous regression formulation \cite{sanchez12} is a special case. Crucially, our method is now formulated by means of a \textbf{full covariance matrix capturing real statistics} of how faces vary between consecutive frames rather than on the shape model eigenvalues. This makes our method particularly suitable for the task of tracking, something the original formulation cannot deal with.
\item We incorporate continuous regression in the Cascaded Regression framework (coined Cascaded Continuous Regression, or \textbf{CCR}) and demonstrate its performance is equivalent to sampling-based cascaded regression. 
\item We derive the \textbf{incremental learning for continuous regression}, and show that it \textbf{is an order of magnitude faster} than its standard incremental SDM counterpart.
\item We evaluate the incremental Cascaded Continuous Regression (\textbf{iCCR}) on the 300VW data set \cite{shen15} and show the importance of incremental learning in achieving state-of-the-art performance, especially for the case of very challenging tracking sequences. 
\end{itemize}

\subsection{Prior Work on Face Alignment}

\noindent Facial landmark tracking methods have often been adaptations of facial landmark detection methods. For example, Active Appearance Models (AAM) \cite{cootes01,matthews04}, Constrained Local Models (CLM) \cite{saragih11} or the Supervised Descent Method (SDM) \cite{xiong13} were all presented as detection algorithms. It is thus natural to group facial landmark tracking algorithms in the same way as the detection algorithms, i.e. splitting them into discriminative and generative methods \cite{asthana14}. 

On the generative side, AAMs have often been used for tracking. Since the model fitting relies on gradient descent, it suffices to start the fitting from the last solution\footnote{Further ``implementation tricks'' can be found in \cite{tresadern12}, which provides a very detailed account of how to optimise an AAM tracker}. Tracking is particularly useful to AAMs since they are considered to have frequent local minima and a small basin of attraction, making it important that the initial shape is close to the ground truth. AAMs have further been regarded as very reliable for person specific tracking, but not for generic tracking (i.e., tracking faces unseen during training) \cite{gross05}. Recently \cite{tzimiropoulos13} showed however that an improved optimisation procedure and the use of in-the-wild images for training can lead to well-behaving person independent AAM. Eliminating the piecewise-affine representation and adopting a part-based model led to the Gauss-Newton Deformable Part Model (GN-DPM) \cite{tzimiropoulos14}, which is the AAM state of the art.

Historically, discriminative methods relied on the training of local classifier-based models of appearance, with the local responses being then constrained by a shape model \cite{cootes95,cristinacce06,saragih11}. These algorithms can be grouped into what is called the Constrained Local Models (CLM) framework \cite{saragih11}. However, the appearance of discriminative regression-based models quickly transformed the state-of-the-art for face alignment. Discriminative regressors were initially used within the CLM framework substituting classifiers, showing improved performance \cite{cootes12,valstar10_CVPR}. However, the most important contributions came with the adoption of cascaded regression \cite{dollar10} and direct estimation of the full face shape rather than first obtaining local estimates \cite{cao14,xiong13}. Successive works have further shown the impressive efficiency \cite{ren14,kazemi14} and reliable performance \cite{yan13,tzimiropoulos15} of face alignment algorithms using cascaded regression. However, how to best exploit discriminative cascaded regression for tracking and, in particular, how to best integrate incremental learning, is still an open problem.

\section{Linear Regression Models for Face Alignment}
\label{sec:prior_work}

\noindent In this section we revise the preliminary concepts over which we build our method. In particular, we describe the methods most closely related to ours, to wit the incremental supervised descent method \cite{asthana14} and the continuous regressor \cite{sanchez12}, and motivate our work by highlighting their limitations.

\subsection{Linear Regression}

\noindent A face image is represented by $\I$, and a face shape is a $n \times 2$ matrix describing the location of the $n$ landmarks considered. A shape is parametrised through a Point Distribution Model (PDM) \cite{cootes2004}. In a PDM, a shape $\s$ is parametrised in terms of $\p = [\q,\c] \in \Real^{m}$, where $\q\in\Real^{4}$ represents the rigid parameters and $\c$ represents the flexible shape parameters, so that $\s = t_\q( \s_0 + \B_s \c )$, where $t$ is a Procrustes transformation parametrised by $\q$. $\B_s \in \Real^{ 2n \times m}$ and $\s_0\in \Real^{ 2n}$ are learned during training and represent the linear subspace of flexible shape variations. We will sometimes use an abuse of notation by referring treating shape $\s$ also as function $\s( \p )$. We also define $\x = f( \I , \p ) \in \Real^d $ as the feature vector representing shape $\s( \p )$. An asterisk represents the ground truth, e.g., $\s_j^*$ is the ground truth shape for image $j$. 

Given a test image $\I$, and a current shape prediction $\s( \p^* + \delta \p )$, the goal of Linear Regression for face alignment is to find a mapping matrix $\R \in \Real^{m \times d}$ able to infer $\delta \p$, the increment taking directly to the ground truth, from $f(\I,\p^* + \delta \p)$. By using $M$ training images, and $K$ perturbations per image, the mapping matrix $\R$ is typically learned by minimising the following expression w.r.t. $\R$:

\begin{equation}
\label{eq:standard_LinearReg_problem}
\sum_{j=1}^M \sum_{k=1}^K \| \delta \p_{j,k} - \R f( \I_j , \p^*_j + \delta \p_{j,k} ) \|_2^2,
\end{equation}

\noindent where the bias term is implicitly included by appending a 1 to the feature vector\footnote{It is in practice beneficial to include a regularisation term, although we omit it for simplicity. All of the derivations in this paper hold however for ridge regression}. 

In order to produce $K$ perturbed shapes $\s( \p_j^* + \delta \p_{j,k} )$ per image, it suffices to draw the perturbations from an adequate distribution, ideally capturing the statistics of the perturbations encountered at test time. For example, during detection, the distribution should capture the statistics of the errors made by using the face detection bounding box to provide a shape estimation. 

The minimisation in Eq.~\ref{eq:standard_LinearReg_problem} has a closed-form solution. Given $M$ images and $K$ perturbed shapes per training image, let $\X \in \Real^{d \times KM}$ and $\Y \in \Real^{2n \times KM}$ represent the matrices containing in its columns the input feature vectors and the target output $\delta \p_{j,k}$ respectively. Then, the optimal regressor $\R$ can be computed as:

\begin{equation}
\label{eq:linear_reg}
\R = \Y \X^T \left( \X \X^T \right)^{-1}.
\end{equation}

Given a test shape $\s( \p )$, the predicted shape is computed as $\s( \p - \R f(\I,\p) )$.

\subsection{Continuous Regression}

\noindent Continuous Regression (CR) \cite{sanchez12} is an alternative solution to the problem of linear regression for face alignment. The main idea of Continuous Regression is to treat $\delta \p$ as a continuous variable and to use \textit{all samples} within some finite limits, instead of sampling a handful of perturbations per image. That is to say, the problem is formulated in terms of finite integrals as:  

\begin{equation}
\label{eq:old_continuous}
\min_\R \sum_{j=1}^M \int_{-r_1 \sqrt{\lambda_1}}^{r_1 \sqrt{\lambda_1}} \dots \int_{-r_{|\c|} \sqrt{\lambda_{|\c|}}}^{r_{|\c|} \sqrt{\lambda_{|\c|}}}\|\delta \c - \R f(\I_j, \c_j^*+ \delta \c)\|_2^2 d \delta \c,
\end{equation}

\noindent where $\lambda_i$ is the eigenvalue associated to the $i$-th flexible parameter of the PDM, $|\c|$ represent the number of flexible parameters, and $r_i$ is a parameter determining the number of standard deviations considered in the integral. 

Unfortunately, this formulation does not have a closed-form solution. However, it is possible to solve it approximately in a very efficient manner by using a first order Taylor expansion of the loss function. Following the derivations in \cite{sanchez12}, we denote $\J_j^*$ as the Jacobian of the image features with respect to the shape parameters evaluated at the ground truth $\p_j^*$, which can be calculated simply as $\J^*_j = \frac{\partial f(\I_j, \s)}{\partial \s} \frac{\partial \s}{ \partial \p} \vert_{(\p = \p_j^*)}$. A solution to Eq.~\ref{eq:old_continuous} can then be written as:

\begin{equation}
\label{eq:old_continuous_sol}
\R( \boldsymbol{\bf r}) = \Sig(\boldsymbol{\bf r}) (\sum_{j=1}^M {\J_j^*}^T ) \left(  \sum_{j=1}^M \x_j^* {\x_j^*}^T +  \J_j^* \Sig(\boldsymbol{\bf r}) \ {\J_j^*}^T \right)^{-1},
\end{equation}

\noindent where $\Sig(\boldsymbol{\bf r})$ is a diagonal matrix whose $i$-th entries are defined as $\frac{1}{3}r_i^2 \lambda_i$. CR formulated in this manner has the following practical limitations:

\begin{enumerate}
\item It does not account for correlations within the perturbations. This corresponds to using a fixed (not data-driven) diagonal covariance to model the space of shape perturbations, which is a harmful oversimplification.
\item Because of 1, it is not possible to incorporate CR within the popular cascaded regression framework in an effective manner. 
\item Derivatives are computed over image pixels, so more robust features, e.g., HOG or SIFT, are not used.
\item The CR can only account for the flexible parameters, as the integral limits are defined in terms of the eigenvalues of the PDM's PCA space.
\end{enumerate}

In Section \ref{ssec:continuous_regression} we will solve all of these shortcomings, showing that it is possible to formulate the cascaded continuous regression and that, in fact, its performance is equivalent to the SDM.

\subsection{Supervised Descent Method}

\noindent The main limitation of using a single Linear Regressor to predict the ground truth shape is that the training needs to account for too much intra-class variation. That is, it is hard for a single regressor to be simultaneously accurate and robust. To solve this, \cite{cao12} successfully adapted the cascaded regression of framework of Doll{\'a}r et al. \cite{dollar10} to  face alignment. However, the most widely-used form of face alignment is the SDM \cite{xiong13}, which is a cascaded linear regression algorithm. 

At \textbf{test time}, the SDM takes an input $\s( \p^{(0)} )$, and then for a fixed number of iterations computes $\x^{(i)}=f(\I,\p^{(i)})$ and $\p^{(i+1)} = \p^{(i)} - \R^{(i)} \x^{(i)}$. The key idea is to use a different regressor $\R^{(i)}$ for each iteration. The input to the \textbf{training} algorithm is a set of images $\I_j$ and corresponding perturbed shapes $\p_{j,k}^{(0)}$. The training set $i$ is defined as $\X^{(i)}=\{\x_{j,k}^{(i)}\}_{j=1:M,k=1:K}$, with $\x_{j,k}^{(i)}=f( \I_j,\p_{j,k}^{(i)})$, and $\Y^{(i)}=\{\y_{j,k}^{(i)}\}_{j=1:M,k=1:K}$, with $\y_{j,k}^{(i)}=\p_{j,k}^{(i)}-\p_j^*$. Then regressor $i$ is computed using Eq.~\ref{eq:linear_reg} on training set $i$, and a new training set $\{\X^{(i+1)},\Y^{(i+1)}\}$ is created using the shape parameters $\p_{j,k}^{(i+1)}=\p_{j,k}^{(i)} - \R^{(i)}\x_{j,k}^{(i)}$.

\subsection{Incremental Learning for SDM}
\label{ssec:inc_learn_SDM}

\noindent Incremental versions of the SDM have been proposed by both Xiong \& De la Torre \cite{xiong14} and Asthana et al. \cite{asthana14}. The latter proposed the \emph{parallel SDM}, a modification of the original SDM which facilitates the incremental update of the regressors. More specifically, they proposed to alter the SDM training procedure by modelling $\{ \p_{j,k}^{(i)} - \R^{(i)}\x_{j,k}^{(i)} \}_{j,k}$ as a Normal distribution $\mathcal{N}( \bmu^{(i)},\Sig^{(i)})$, allowing training shape parameters to be sampled for the next level of the cascade as:

\begin{equation}
\label{eq:parSDM_sampling}
\p_{j,k}^{(i+1)} \sim \mathcal{N}(\p_j^{*}+\bmu^{(i)},\Sig^{(i)})
\end{equation}

Once the parallel SDM is defined, its incremental extension is immediately found. Without loss of generality, we assume that the regressors are updated in an on-line manner, i.e., the information added is extracted from the fitting of the last frame. We thus define $\mathcal{S} =\{\I_j , \{\p_{j,k}\}_{k=1}^K \}$, arrange the matrices $\X_\mathcal{S}$ and $\Y_\mathcal{S}$ accordingly, and define the shorthand $\V_\T = \left( \X_\T \X_\T^T \right)^{-1}$, leading to the following update rules \cite{asthana14}:

\begin{eqnarray}
\R_{\T \cup \mathcal{S}} & = & \R_\T - \R_\T \Q + \Y_{\mathcal{S}}\X_{\mathcal{S}}^T\V_{\mathcal{T}\cup \mathcal{S}} \label{eq:incremental_discrete_eq1}\\
\Q & = & \X_{\mathcal{S}} \U \X_{\mathcal{S}}^T\V_{\mathcal{T}} \label{eq:incremental_discrete_eq2}\\
\U & = & \left( \mathbb{I}_K + \X_{\mathcal{S}}^T \V_{\mathcal{T}} \X_{\mathcal{S}} \right)^{-1}  \label{eq:incremental_discrete_eq3}\\
\V_{\mathcal{T} \cup \mathcal{S}} & = & \V_{\mathcal{T}}- \V_{\mathcal{T}}\Q \label{eq:incremental_discrete_eq4}
\end{eqnarray}

\noindent where $\mathbb{I}_K$ is the $K$-dimensional identity matrix. 

The cost for these incremental updates is dominated by the multiplication $\V_\T \Q$, where both matrices have dimensionality $d \times d$, which has a computational complexity of $\oO(d^3)$. Since $d$ is high-dimensional ($>1000$), the cost of updating the models becomes prohibitive for real-time performance. Once real time is abandoned, offline techniques that do not analyse every frame in a sequential manner can be used for fitting, e.g., \cite{sagonas14}. We provide a full analysis of the computational complexity in Section~\ref{sec:computational_complexity}.

\section{Incremental Cascaded Continuous Regression (iCCR)}

\noindent In this section we describe the proposed Incremental Cascaded Continuous Regression, which to the best of our knowledge is the first cascaded regression tracker with \textbf{real-time incremental learning} capabilities. To do so, we first extend the Continuous Regression framework into a fully fledged cascaded regression algorithm capable of performance on par with the SDM (see Sections~\ref{ssec:continuous_regression} and \ref{ssec:cascaded_continuous_regression}). Then, we derive the incremental learning update rules within our Cascaded Continuous Regression formulation (see Section~\ref{ssec:incremental_learning_update_rules}). We will show in Section~\ref{sec:computational_complexity} that our newly-derived formulas have complexity of one order of magnitude less than previous incremental update formulations.

\subsection{Continuous Regression Revisited}
\label{ssec:continuous_regression}

\noindent We first modify the original formulation of Continuous Regression. In particular, we add a ``data term'', which is tasked with encoding the probability of a certain perturbed shape, allowing for the modelling of correlations in the shape dimensions. Plainly speaking, the previous formulation assumed an i.i.d. uniform sampling distribution. We instead propose using a data-driven full covariance distribution, resulting in regressors that model the test-time scenario much better. In particular, we can see the loss function to be optimised as:

\begin{equation}
\label{eq:train_problem_def}
\argmin_{\R} \sum_{j=1}^M \int_{\delta \p} p( \delta \p ) \| \delta \p - \R f( \I_j, \p^*_j+\delta \p ) \|_2^2 d \delta \p.
\end{equation}

\begin{figure}[t!]
\begin{center}
\includegraphics[width=0.85\columnwidth]{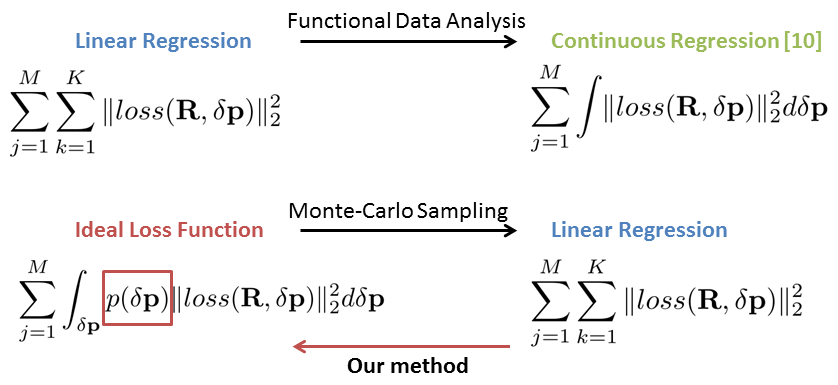}
\end{center}
\caption{Main difference between original Continuous Regression \cite{sanchez12} and our method. \label{fig:differences}}
\end{figure}

It is interesting to note that this equation appears in \cite{xiong13}, where the SDM equations are interpreted as a MCMC sampling-based approximation of this equation. Contrariwise, the Continuous Regression proposes to use a different approximation based on a first-order Taylor approximation of the \textit{ideal loss function} defined in Eq.~\ref{eq:train_problem_def}. However, the Continuous Regression proposed in \cite{sanchez12} extends the Functional Data Analysis \cite{brookes11} framework to the imaging domain, without considering any possible data correlation. Instead, the ``data term'' in Eq.~\ref{eq:train_problem_def} (which defines how the data is sampled in the MCMC approach), will serve to correlate the different dimensions in the Continuous Regression. That is to say, the ``data term" does not play the role of how samples are taken, but rather helps to find an analytical solution in which dimensions can be correlated. These differences are illustrated in Figure~\ref{fig:differences}. 

The first-order approximation of the feature vector is given by:

\begin{equation}
f(\I_j,\p_j^*+\delta \p ) \approx f(\I_j,\p_j^*) + \J_j^* \delta \p
\label{eq:taylor_feat}
\end{equation}

\noindent where $\J_j^*$ is the Jacobian of the feature representation of image $\I_j$ at $\p_j^*$. While \cite{sanchez12} used a pixel-based representation, the Jacobian under an arbitrary representation can be computed empirically as:

\begin{equation}
\J_x = \frac{\partial f( \I,\s)}{\partial x} \approx \frac{f( \I, [\s_x + \Delta x, \s_y] ) - f( \I, [\s_x - \Delta x, \s_y] )}{2 \Delta x}
\end{equation}

\noindent where $\s_x$ are the $x$ coordinates of shape $\s$, and $\s_x + \Delta x$ indicates that $\Delta x$ is added to each element of $\s_x$ (in practice, $\Delta x$ is the smallest possible, 1 pixel). $\J_y$ can be computed similarly. Then $\J_j^* = [ \J_x,\J_y] \frac{\partial\s}{ \partial \p_j^*}$. Eq.~\ref{eq:train_problem_def} has a closed form solution as\footnote{A full mathematical derivation is included in the Supplementary Material}:

\begin{multline}
\label{eq:closed_form_cont}
\R_\T = \left( \sum_{j=1}^M  \bmu {\x_j^*}^T+ (\Sig + \bmu \bmu^T){\J_j^*}^T \right) \cdot  \\
\left( \sum_{j=1}^M \x_j^*{\x_j^*}^T+2\x_j^*\bmu^T{\J_j^*}^T + \J_j^*(\Sig+ \bmu\bmu^T){\J_j^*}^T \right) ^{-1}
\end{multline}

\noindent where $\bmu$ and $\Sig$ are the mean and covariance of the data term, $p(\delta \p)$.

Finally, we can see that Eq.~\ref{eq:closed_form_cont} can be expressed in a more compact form. Let us first define the following shorthand notation: ${\bf A} = [ \bmu , \Sig + \bmu \bmu^T ]$, ${\bf B} = \bigl( \begin{smallmatrix} 1 & \bmu^T \\ \bmu & \Sig + \bmu \bmu^T \end{smallmatrix} \bigr)$, $\D_j^* = [ \x_j^*, \J_j^*]$ and $\bar{\D}_\mathcal{T}^*=\left[ \D_1^*,\ldots, \D_M^* \right] $. Then:

\begin{equation}
\label{eq:compact_form_cont}
\R_\T = \A \left( \sum_{j=1}^{M} \D_j^* \right)^T \left(\bar{\D}_\mathcal{T}^* \hat{\B} (\bar{\D}^*_\mathcal{T})^T \right)^{-1}
\end{equation}

\noindent where $\hat{\bf B} = {\bf B} \otimes {\mathbb{I}}_M$. Through this arrangement, the parallels with the sampling-based regression formula are clear (see Eq.~\ref{eq:linear_reg}). 

It is interesting that, while the standard linear regression formulation needs to sample perturbed shapes from a distribution, the Continuous Regression training formulation only needs to extract the features and the Jacobians on the ground-truth locations. This means that once these features are obtained, re-training a new model under a different distribution takes seconds, as it only requires the computation of Eq.~\ref{eq:compact_form_cont}.

\subsection{Cascaded Continuous Regression (CCR)}
\label{ssec:cascaded_continuous_regression}

\noindent Now that we have introduced a new formulation with the Continuous Regression capable of incorporating a data term, it is straightforward to extend the CR into the cascade regression formulation: we take the distribution in Equation~\ref{eq:parSDM_sampling} as the \textit{data term} in Eq.~\ref{eq:train_problem_def}. 

One might argue that due to the first-order Taylor approximation required to solve Equation~\ref{eq:train_problem_def}, CCR might not work as well as the SDM. One of the main experimental contributions of this paper is to show that in reality this is not the case: in fact CCR and SDM have equivalent performance (see Section~\ref{sec:experimental_results}). This is important because, contrary to previous works on Cascaded Regression, incremental learning within CCR allows for real time performance.

\subsection{Incremental Learning Update Rules for CCR}
\label{ssec:incremental_learning_update_rules}

\noindent Once frame $j$ is tracked, the incremental learning step updates the existing training set $\mathcal{T}$ with $\S=\{\I_j,\hat{\p}_{j}\}$, where $\hat{\p}_{j}$ denotes the predicted shape parameters for frame $j$. Note that in this case $\S$ consists of only one example compared to $K$ examples in the incremental SDM case. 

The update process consists of computing matrix $\D_j$, which stores the feature vector and its Jacobian at $\hat{\p}_{j}$ and then, using the shorthand notation $\V_\mathcal{T} = \bar{\D}_\mathcal{T}^* \hat{\B}(\bar{\D}_\mathcal{T}^*)^T$, updating continuous regressor as:

\begin{equation}
\R_{\mathcal{T} \cup \mathcal{S}} = \A \left(  \sum_{j=1}^{M} \D_j^* +  \D_\mathcal{S}^* \right)^T \left( \V_{\mathcal{T} \cup \mathcal{S}}\right)^{-1}
\end{equation}

In order to avoid the expensive re-computation of $\V_\mathcal{T}^{-1}$, it suffices to update its value using the Woodbury identity \cite{brookes11}:

\begin{equation}
\label{eq:closed_form_icont}
{\V_{\mathcal{T} \cup \mathcal{S}}}^{-1} = {\V_{\mathcal{T}}}^{-1} - 
 {\V_{\mathcal{T}}}^{-1} \D_{\mathcal{S}}^*\left( \B^{-1}+{\D_{\mathcal{S}}^*}^T {\V_{\mathcal{T}}}^{-1} \D_{\mathcal{S}}^* \right)^{-1} {\D_{\mathcal{S}}^*}^T{\V_{\mathcal{T}}}^{-1}
\end{equation}

Note that $\D_\mathcal{S}^* \in \Real^{d \times (m+1)}$, where  $m$ accounts for the number of shape parameters. We can see that computing Eq.~\ref{eq:closed_form_icont} requires computing first ${\D_{\mathcal{S}}^*}^T{\V_{\mathcal{T}}}^{-1}$, which is $\oO( m d^2)$. This is a central result of this paper, and reflects a property previously unknown. We will examine in Section~\ref{sec:computational_complexity} its practical implications in terms of real-time capabilities.

\section{Computational Complexity}
\label{sec:computational_complexity}

\noindent In this section we first detail the computational complexity of the proposed iCCR, and show that it is real-time capable. Then, we compare its cost with that of incremental SDM, showing that our update rules are an order of magnitude faster.

\textbf{iCCR update complexity:} Let us note the computational cost of the feature extraction as $\oO(q)$. The update only requires the computation of the feature vector at the ground truth, and in two adjacent locations to compute the Jacobian, thus resulting in $\oO(3q)$ complexity. Interestingly, this is independent from the number of cascade levels.

Then, the update equation (Eq.~\ref{eq:closed_form_icont}), has a complexity dominated by the operation ${\D_{\mathcal{S}}}^T{\V^{\mathcal{C}}_{\mathcal{T}}}^{-1}$, which has a cost of $\mathcal{O}(d^2 m )$. It is interesting to note that $\B^{-1}+{\D_{\mathcal{S}}}^T {\V^{\mathcal{C}}_{\mathcal{T}}}^{-1} \D_{\mathcal{S}}$ is a matrix of size $(m+1) \times (m+1)$ and thus its inversion is extremely efficient. The detailed cost of the incremental update is:

\begin{equation}
\label{eq:cost_cont}
\oO( 3m d^2) + \oO(3m^2 d) + \oO(m^3).
\end{equation}

\textbf{Incremental SDM update complexity:} Incremental learning for SDM requires sampling at each level of the cascade. The cost per cascade level is $\mathcal{O}( q K )$, where $K$ denotes the number of samples. Thus, for $L$ cascade levels the total cost of sampling is $\oO(LKq)$. The cost of the incremental update equations (Eqs.~(\ref{eq:incremental_discrete_eq1}-\ref{eq:incremental_discrete_eq4})), is in this case dominated by the multiplication $\V_{\mathcal{T}}\Q$, which is $\oO(d^3)$. The detailed computational cost is:

\begin{equation}
\label{eq:cost_sampled}
\oO(d^3) + \oO( (3m+k) d^2) + \oO((2K^2 + m k) d) + \oO(K^3).
\end{equation}

\textbf{Detailed comparison and timing:} One advantage of iCCR comes from the much lower number of feature computations, being as low as 3 vs. the $LK$ computations required for incremental SDM. However, the main difference is the $\mathcal{O}(d^3)$ complexity of the regressor update equation for the incremental SDM compared to $\mathcal{O}(d^2m)$ for the iCCR. In our case, $d=2000$, while $m=24$. The feature dimensionality results from performing PCA over the feature space, which is a standard procedure for SDM. Note that if we avoided the use of PCA, the complexity comparison would be even more in our favour. A detailed summary of the operations required by both algorithms, together with their computational complexity and the execution time on our computer are given in Algorithm~\ref{algo_iccr}. Note that $\oO(D)$ is the cost of projecting the output vector into the PCA space. Note as well that for incremental SDM, the ``Sampling and Feature extraction'' step is repeated $L$ times.

\RestyleAlgo{boxruled}
\begin{algorithm}[t!]
\SetKwFunction{Union}{Union}\SetKwFunction{FindCompress}{FindCompress}
\SetKwInOut{Precompute}{precompute}\SetKwInOut{Output}{output}
\textit{iCCR update (Total: 72 ms.):} \\
\Precompute{ Feature and Jacobian extraction \, : $ \langle \oO(3q) : 9 \mbox{ ms.}\rangle$}
\BlankLine
\For{$i\leftarrow 1$ \KwTo $L=3$ cascade levels}{
PCA Projection  \hspace{17pt}        \, : $ \langle \oO\left(D m\right) : 6 \mbox{ ms.}\rangle$ \;
Update $\R$ (Eq.~\ref{eq:closed_form_icont}) \, : $ \langle \oO(m d^2) : 15 \mbox{ ms.}\rangle$ \;
}
\noindent\rule{11cm}{0.4pt}
\BlankLine
\textit{iSDM \cite{asthana14} update (Total: 705 ms.):} \\
\For{$i\leftarrow 1$ \KwTo $L=3$ cascade levels}{
Sampling and Feature extraction \, : $ \langle \oO(Kq) : 30 \mbox{ ms.}\rangle$ \;
PCA Projection  \hspace{17pt}        \, : $ \langle \oO(DK) : 5 \mbox{ ms.}\rangle$ \;
Update $\R$ (Eqs.~\ref{eq:incremental_discrete_eq1}-\ref{eq:incremental_discrete_eq4}) \hspace{3pt} \, : $ \langle \oO(d^3) : 200 \mbox{ ms.}\rangle$ \;
}
\caption{Computational costs for iCCR and incremental SDM \cite{asthana14} updates}\label{algo_iccr}
\end{algorithm}

\section{Experimental results}
\label{sec:experimental_results}
\noindent This section describes the experimental results. First, we empirically demonstrate the performance of CCR is equivalent to SDM. In order to do so, we assess both methods under the same settings, avoiding artefacts to appear, such as face detection accuracy. We follow the VOT Challenge protocol \cite{kristan15}.  Then, we develop a fully automated system, and we evaluate both the CCR and iCCR in the same settings as the 300VW, and show that our fully automated system achieves state of the art results, illustrating the benefit of incremental learning to achieve it.
\subsection{Experimental set-up}
\label{ssec:Experimental_set_up}

\noindent \textbf{Training Data:} We use data from different datasets of static images to construct our training set. Specifically, we use Helen \cite{le12}, LFPW \cite{belhumeur11}, AFW \cite{zhu12}, IBUG \cite{sagonas13}, and a subset of MultiPIE \cite{gross10}. The training set comprises $\sim$7000 images. We have used the facial landmark annotations provided by the 300 faces in the wild challenge \cite{sagonas13}, as they offer consistency across datasets. The \textit{statistics} are computed across the training sequences, by computing the differences of ground-truth shape parameters between consecutive frames. Given the easiness of the training set with respect to the test set, we also included differences of several frames ahead. This way, higher displacements are also captured. 

\noindent \textbf{Features:} We use the SIFT \cite{lowe04} implementation provided by Xiong \& De la Torre \cite{xiong13}. We apply PCA on the output, retaining 2000 dimensions. We apply the same PCA to all of the methods, computed during our SDM training.

\noindent \textbf{Test Data:} All the methods are evaluated on the test partition of the 300 Videos in the Wild challenge (300VW) \cite{shen15}. The 300VW is the only publicly-available large-scale dataset for facial landmark tracking. Its test partition has been divided into categories 1, 2 and 3, intended to represent increasingly unconstrained scenarios. In particular, category 3 contains videos captured in totally unconstrained scenarios. The ground truth has been created in a semi-supervised manner using two different methods \cite{tzimiropoulos15,chrysos15}. 

\noindent \textbf{Error measure:} To compute the error for a specific frame, we use the error measure defined in the 300VW challenge \cite{shen15}. The error is computed by dividing the average point-to-point Euclidean error by the inter-ocular distance, understood as the distance between the two outer eye corners. 

\subsection{CCR vs SDM}
\label{ssec:CCRvsSDM}

\noindent In order to demonstrate the performance capability of our CCR method against SDM, we followed the protocol established by the Visual Object Tracking (VOT) Challenge organisers for evaluating the submitted tracking methods \cite{kristan15}. Specifically, if the tracker error exceeds a certain threshold (0.1 in our case, which is a common definition of alignment failure), we proceed by re-initialising the tracker. In this case, the starting point will be the ground truth of the previous frame. This protocol is adopted to avoid the pernicious influence on our comparison of some early large failure from which the tracker is not able to recover, which would mean that successive frames would yield a very large error. Results are shown in Fig.~\ref{fig:ccrvssdm} (\textbf{Left}). We show that the CCR and the SDM provide similar performance, thus ensuring that the CCR is a good starting point for developing an incremental learning algorithm. It is possible to see from the results shown in Fig.~\ref{fig:ccrvssdm} that the CCR compares better and even sometimes surpasses the SDM on the lower levels of the error, while the SDM systematically provides a gain for larger errors with respect to the CCR. This is likely due to the use of first-order Taylor approximation, which means that larger displacements are less accurately approximated. Instead, the use of \textit{infinite} shape perturbations rather than a handful of sampled perturbations compensates this problem for smaller errors, and even sometimes provides some performance improvement.

\subsection{CCR vs iCCR}
\label{ssec:CCRvsiCCR}
\begin{figure}
\begin{center}
\includegraphics[width=0.45\columnwidth]{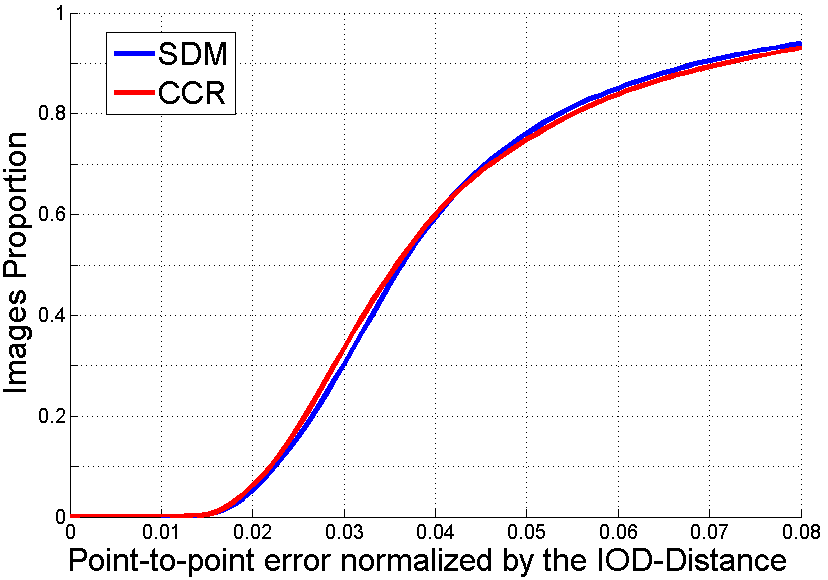} 
\includegraphics[width=0.45\columnwidth]{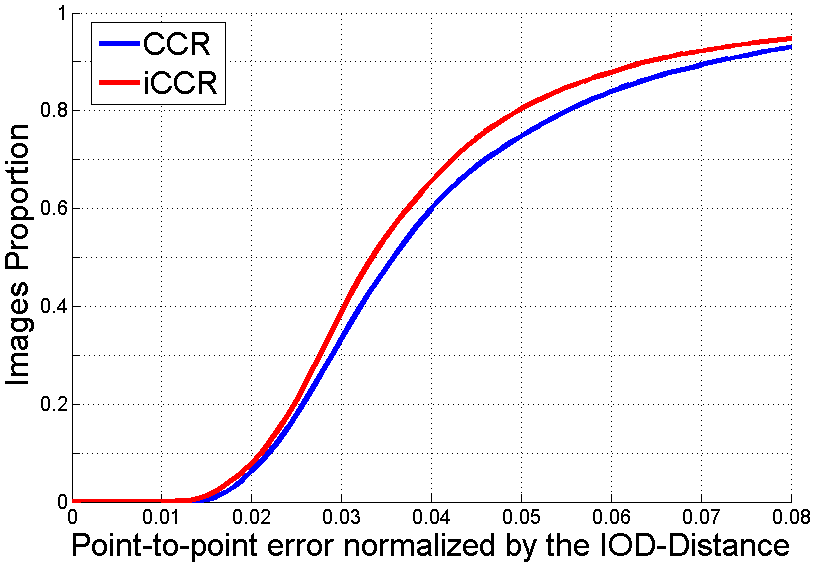}
\end{center}
\caption{\textbf{Left}: Accumulated graph across all three categories for SDM and CCR methods. In both cases, the Area Under the Curve (AUC) is 0.49, meaning that CCR shows better capabilities for lower errors, whereas SDM fits better in higher errors. \textbf{Right}: Accumulated graph across all three categories for CCR and iCCR methods. The contribution of incremental learning is clear. \label{fig:ccrvssdm}}
\end{figure}

\noindent We now show the benefit of incremental learning with respect to generic models. The incremental learning needs to filter frames to decide whether a fitting is suitable or harmful to update the models. That is, in practice, it is beneficial to filter out badly-tracked frames by avoiding performing incremental updates in these cases. We follow \cite{asthana14} and use a linear SVM trained to decide whether a particular fitting is ``correct'', understood as being under a threshold error. Despite its simplicity, this tactic provides a solid performance increase. Results on the test set are shown in Fig.~\ref{fig:ccrvssdm} (\textbf{Right}). 

\subsection{Comparison with state of the art}
\label{ssec:comparisons_with_sota}
\begin{figure}
\begin{center}
\includegraphics[width=0.94\columnwidth]{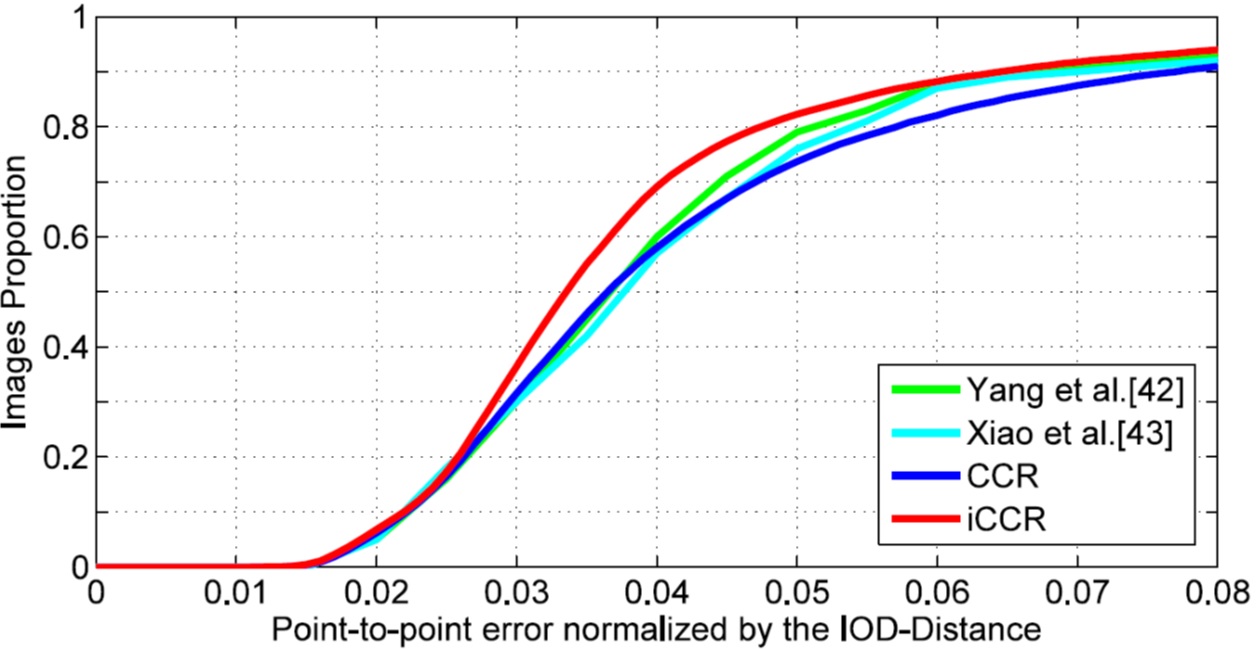}
\end{center}
\caption{Results given by our fully automated system in the most challenging category of the 300VW benchmark. Results are shown for the 49 inner points. The contribution of Incremental Learning in challenging sequences, and in a fully automated system, is even higher. \label{fig:soacomp} }
\end{figure}

\noindent We developed a fully automated system to compare against state of the art methods. Our fully automated system is initialised with a standard SDM \cite{sanchez16}, and an SVM is used to detect whether the tracker gets lost. We assessed both our CCR and iCCR in the most challenging category of the 300VW, consisting of 14 videos recorded in unconstrained settings. For a fair comparison, we have reproduced the challenge settings (a brief description of the challenge and submitted methods can be found in \cite{shen15}). We compare our method against the top two participants \cite{yang15,xiao15}. 
Results are shown in Fig.~\ref{fig:soacomp}. The influence of the incremental learning to achieve state of the art results is clear. Importantly, as shown in the paper, our iCCR allows for real-time implementation. That is to say, our iCCR reports state of the art results whilst working in near real-time, something that could not be achieved by previous works on Cascaded Regression. Code for our fully automated system is available for download at \url{www.cs.nott.ac.uk/~psxes1}.

\section{Conclusion}

\noindent In this article we have proposed a novel facial landmark tracking algorithm that is capable of performing on-line updates of the models through incremental learning. Compared to previous incremental learning methodologies, it can produce much faster incremental updates without compromising on accuracy. This was achieved by firstly extending the Continuous Regression framework \cite{sanchez12}, and then incorporating it into the cascaded regression framework to lead to the CCR method, which we showed provides equivalent performance to the SDM. We then derived the incremental learning update formulas for the CCR, resulting in the iCCR algorithm. We further show the computational complexity of the incremental SDM, demonstrating that iCCR is an order of magnitude simpler computationally. This removes the bottleneck impeding real-time incremental cascaded regression methods, and thus results in the state of the art for real-time face tracking.

\section*{Acknowledgments}

\noindent The work of S{\'a}nchez-Lozano, Martinez and Valstar was supported by the European Union Horizon 2020 research and innovation programme under grant agreement No 645378, ARIA-VALUSPA. The work of S{\'a}nchez-Lozano was also supported by the Vice-Chancellor's Scholarship for Research Excellence provided by the University of Nottingham. The work of Tzimiropoulos was supported in part by the EPSRC project EP/M02153X/1 Facial Deformable Models of Animals. We are also grateful for the given access to the University of Nottingham High Performance Computing Facility, and we would like to thank Jie Shen and Grigoris Chrysos for their insightful help in our tracking evaluation.

\bibliographystyle{splncs}

\newpage
\appendix
\section{ Derivation of CCR }

\noindent As shown in the paper, linear regression aims at minimising the average expected training error, respect to $\R$. The average expected error is formulated as follows: 

\begin{equation}
\label{eq:train_problem_def_appx}
\argmin_{\R} \sum_{j=1}^M \int_{\delta \p} p( \delta \p ) \| \delta \p - \R f( \I_j, \p^*_j+\delta \p ) \|_2^2 d \delta \p,
\end{equation}
where $\p^*_j$ represents the ground truth shape parameters, and $\delta \p$ the parameters displacement. Due to the intractability of the integral, this is typically solved by a MCMC (sampling-based) approximation, in which samples are taken from the distribution $p(\delta \p)$. The continuous regression framework avoids the need to sample by performing the first order Taylor approximation of the function $f$, defined as:

\begin{equation}
f(\I_j,\p_j^*+\delta \p ) \approx f(\I_j,\p_j^*) + \J^*_j \delta \p
\label{eq:taylor_feat}
\end{equation}

\noindent where  $\J^*_j = \frac{\partial f(\I_j, \p)}{\partial \s} \frac{\partial \s}{ \partial \p} \vert_{(\p = \p_j^*)}$, evaluated at $\p = \p_j^*$, is the Jacobian of the feature representation of image $\I_j$, respect to shape parameters $\p$, at $\p_j^*$. Combining this approximation with the integral in Eq.~\ref{eq:train_problem_def_appx} leads to:

\begin{eqnarray}
\int_{\delta \p} p( \delta \p ) \| \delta\p - \R f( \I_j, \p^*_j+\delta \p ) \|_2^2 d \delta \p \nonumber \approx \\ 
\approx \int_{\delta \p} p( \delta \p ) \| \delta \p - \R \left( f(\I_j,\p_j^*) + \J^*_j \delta \p \right)\|_2^2 d \delta \p \nonumber = \\
= \int_{\delta \p} p( \delta \p ) \Big[ \delta\p^T\delta\p - 2 \delta\p^T \R (\x^*_j + \J^*_j\delta\p) + (\x^*_j + \J^*_j \delta \p)^T\R^T\R(\x^*_j + \J^*_j \delta\p) \Big] d \delta \p \hspace{-5pt}
\label{eq:long_deriv_CCR_appx}
\end{eqnarray}
\noindent where recall $\x^*_j$ is the shorthand of $f(\I_j, \p_j^*)$. If we group independent, linear, and quadratic terms, respect to $\delta \p$,we can express Eq.~\ref{eq:long_deriv_CCR_appx} as:
\begin{eqnarray}
\int_{\delta \p} p( \delta \p ) \| \delta \p - \R f( \I_j, \p^*_j+\delta\p ) \|_2^2 d \delta \p \nonumber \approx \\ 
\approx \int_{\delta \p} p( \delta \p ) \big[ \delta \p^T \A \delta \p   + 2\delta\p^T\mathbf{b}_j + {\x^*_j}^T \R^T \R \x^*_j \big] d \delta \p,  
\label{eq:long_deriv_CCR2_appx}
\end{eqnarray}
where $\A_j = ( \mathbb{I} - \R \J^*_j )^T ( \mathbb{I} - \R \J^*_j )$ and $\mathbf{b}_j = {\J^*_j}^T \R^T \R \x^*_j - \R \x^*_j$. Let us assume that $p(\delta \p)$ is parametrised by its mean $\bmu$ and covariance $\Sig$. Then, it follows that:
\begin{eqnarray}
\int_{\delta \p} p(\delta \p) d\delta \p = 1,  \quad \int_{\delta \p} \delta \p p(\delta \p) d\delta \p = \bmu, \nonumber \\ 
\int_{\delta \p} p(\delta \p)\delta \p^T \A \delta \p d\delta \p = Tr(\A \Sig ) + \bmu\A\bmu^T ,
\label{eq:integrals2_appx}
\end{eqnarray}
which means that the expected error, for the $j$-th training example, has a closed-form solution as follows:
\begin{equation}
\int_{\delta \p} p( \delta \p ) \| \delta \p - \R f( \I_j, \p^*_j+\delta \p ) \|_2^2 d \delta \p  \approx Tr( \A_j \Sig ) + \bmu^T \A_j \bmu + 2 \bmu^T \mathbf{b}_j + {\x^*_j}^T \R^T \R \x^*_j .
\label{eq:closed_form_appx}
\end{equation}
Now, $\R$ is obtained after minimising Eq.~\ref{eq:closed_form_appx}, whose derivatives are obtained as follows:

\begin{eqnarray}
\frac{\partial}{\partial \R} Tr( \A_j \Sig ) &=& 2 \R \J^*_j \Sig {\J^*_j}^T - 2 \Sig {\J^*_j}^T \nonumber \\
\frac{\partial}{\partial \R} \bmu^T \A_j \bmu &=& 2 \R \J^*_j \bmu \bmu^T {\J^*_j}^T - 2\bmu \bmu^T {\J^*_j}^T\nonumber \\
\frac{\partial}{\partial \R} 2 \bmu^T \mathbf{b}_j &=& 4 \R \x_j^*\bmu^T{\J_j^*}^T  - 2\bmu {\x^*_j}^T \nonumber \\
\frac{\partial}{\partial \R} {\x^*_j}^T \R^T \R \x^*_j &=& 2 \R \x^*_j {\x^*_j}^T.
\end{eqnarray}
This leads to the solution presented in the paper:
\begin{multline}
\label{eq:closed_form_cont_appx}
\R = \left( \sum_{j=1}^M  \bmu {\x_j^*}^T+ (\Sig + \bmu \bmu^T){\J_j^*}^T \right) \cdot  \\
\left( \sum_{j=1}^M \x_j^*{\x_j^*}^T+2\x_j^*\bmu^T{\J_j^*}^T + \J_j^*(\Sig+ \bmu\bmu^T){\J_j^*}^T \right) ^{-1}
\end{multline}

\section{Generalisation with respect to \cite{sanchez12}}
We can see that our new formulation generalises that of \cite{sanchez12}. More specifically, if we solve Eq.~\ref{eq:train_problem_def} for the non-rigid parameters only, we can define $p(\delta \p)$ to be a uniform distribution defined within the limits  $(-r_i \sqrt{ \lambda_i } , r_i \sqrt{ \lambda_i })$, with $\lambda_i$ the eigenvalue assciated to the $i$-th basis of the PDM, and $r_i$ the number of standard deviations considered for that eigenvalue. In such case, Eq.~\ref{eq:train_problem_def} would be defined, for $k$ non-rigid parameters, as:
\begin{equation}
\label{eq:old_continuous_appx}
\argmin_{\R} \sum_{j=1}^M \int_{-r_1 \sqrt{\lambda_1}}^{r_1 \sqrt{\lambda_1}} \dots \int_{-r_k \sqrt{\lambda_k}}^{r_k \sqrt{\lambda_k}}  \| \delta \p - \R f( \I_j, \p^*_j+\delta \p ) \|_2^2 d \delta \p,
 \end{equation}
which is the problem definition that appeared in \cite{sanchez12}. Moreover, we can see that such a uniform distribution would be parametrised by a zero-mean vector and a diagonal covariance matrix whose entries are $\frac{1}{3}r_i^2 \lambda_i$. In such case, Eq.~\ref{eq:closed_form_cont} would be reduced to the solution presented in \cite{sanchez12}. That is to say, the Continuous Regression presented in \cite{sanchez12} assumed a uniform distribution, without connection to tracking statistics, and no correlation between target dimensions were possible. Instead, our formulation accepts a ``data term", which correlates the target dimensions, and allows for its solution for rigid parameters as well. This ``data term" is crucial to the performance of the CCR.

%

\end{document}